\documentclass[runningheads]{llncs}
\usepackage[T1]{fontenc}
\usepackage{float}
\usepackage{caption}
\usepackage{subfig}
\usepackage{graphicx}
\usepackage{multirow}
\usepackage{bm}
\usepackage{amsfonts}
\usepackage{amsmath}
\usepackage{algorithm2e}
\usepackage{svg}
\usepackage{booktabs}
\usepackage{adjustbox}
\usepackage[title]{appendix}
\usepackage{hyperref}       
\usepackage{xcolor}

\hypersetup{
    colorlinks,
    linkcolor={blue!50!black},
    citecolor={blue!50!black},
    urlcolor={blue!50!black}
}

\makeatletter
\newcommand{\@chapapp}{\relax}%
\makeatother

\usepackage{graphicx}
\begin{document}
\title{Image To Tree with Recursive Prompting}
%
%
%
%

%
\author{James Batten\inst{1,2}\and
Matthew Sinclair\inst{1,2} \and
Ben Glocker\inst{1,2} \and
Michiel Schaap\inst{1,2}}

\institute{Imperial College London \and HeartFlow, Inc.}

\maketitle              
\begin{abstract}
Extracting complex structures from grid-based data is a common key step in automated medical image analysis. The conventional solution to recovering tree-structured geometries typically involves computing the minimal cost path through intermediate representations derived from segmentation masks. However, this methodology has significant limitations in the context of projective imaging of tree-structured 3D anatomical data such as coronary arteries, since there are often overlapping branches in the 2D projection. In this work, we propose a novel approach to predicting tree connectivity structure which reformulates the task as an optimization problem over individual steps of a recursive process. We design and train a two-stage model which leverages the UNet and Transformer architectures and introduces an image-based prompting technique. Our proposed method achieves compelling results on a pair of synthetic datasets, and outperforms a shortest-path baseline.

\end{abstract}
\section{Introduction}

Extracting centerline trees from medical images is challenging in projective modalities such as Coronary X-ray Angiography due to overlapping branches in the 2D image. \emph{Tree connectivity structure} extraction is a more restricted problem which involves detecting the set of topologically significant keypoints (such as root, bifurcation and leaf nodes) in the image domain and predicting the directed edges which connect these together to form a tree. We focus on this important subproblem and propose a novel method which achieves favorable performance on two large-scale synthetic datasets compared to a shortest-path baseline.

Our novel two-stage neural network model, named I2TRP, leverages the UNet \cite{Ronneberger2015-tg} and Transformer \cite{Vaswani2017-ez} architectures in order to perform tree connectivity structure extraction in 2D images. The I2TRP model employs an image-based prompting technique and decomposes the tree decoding problem into multiple recursive steps. Formulating the task as an optimization problem over individual steps of a recursive process enables a fully-supervised training strategy while removing the need for more complex end-to-end optimization. Furthermore, we introduce a simple yet effective method to stochastically sample trees from our model, and merge these into a final prediction in order to improve performance. We evaluate our method on two large synthetic datasets and explore the effects of different sampling parameters.

\section{Related Work}

\textbf{Vessel extraction.} The extraction of curvilinear centerline trees is a common task for the analysis of vascular data.  Many classical techniques apply minimal-path algorithms to acquire the curve \cite{Cohen1997-si,Benmansour2011-to,Li2006-ct}. Improvements can be achieved by predicting a dense centerline distance map and subsequently extracting the minimal path \cite{Guo2019-kn}, and decoding topological feature vectors which are then used to inform the tree reconstruction \cite{Keshwani2020-qs}. Other approaches use formal graph-based techniques on skeletonised binary segmentations \cite{Chapman2015-ou}. A more recent method predicts skeletonisations of manual segmentations using convolutional and recurrent architectures \cite{Bates2017-wo}.

\textbf{Set Prediction.} Image-to-Tree prediction can be viewed as an extension to \emph{set prediction}, since the underlying nodes of the tree are set-structured. While object detection is a mature field of research \cite{He2017-pu}, recent approaches have tackled set prediction from novel angles such as end-to-end modeling through the use of permutation-invariant loss functions \cite{Kosiorek2020-ve}. An alternative method, particularly relevant to this paper, discards the end-to-end paradigm in favour of sequence-based modeling \cite{Chen2021-js}, and tackles the set prediction problem by training a neural network to model single steps of a \emph{sequential} process. Our proposed method leverages this intuition by modeling single steps of a \emph{recursive} process.

\textbf{Graph Generation.} Another perspective on tree connectivity structure extraction is to consider it as a subclass of graph generation. In \cite{Jin2018-vq}, the authors propose leveraging a tree-structured intermediate representation in order to support the decoding of the graph-structured molecule. Other approaches in the field of generative molecular modeling include \cite{You2018-fr} which proposes a policy network implemented using graph convolutions, and models the molecular graph decoding as a Markov decision process.

\section{Proposed Method: I2TRP}

\begin{figure}[h]
\centering 
\includegraphics[width=0.8\hsize]{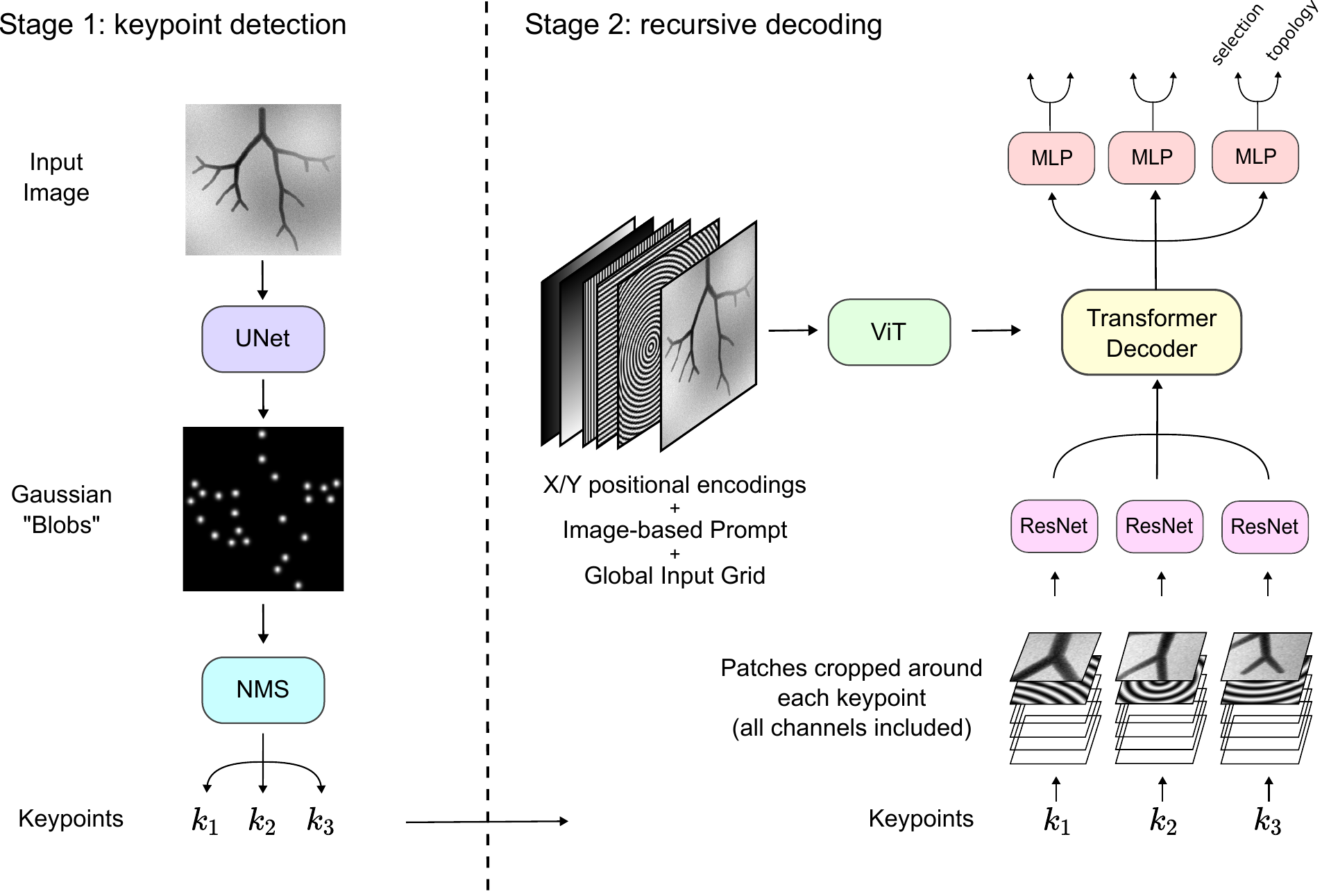}
\captionsetup{justification=centering,margin=0.5cm}
\caption{Overview of the two stages of the I2TRP model. Stage 1 on the left: keypoint detection. Stage 2 on the right: recursive decoding}
\label{fig:model_002}
\end{figure}

The proposed I2TRP (``Image To Tree with Recursive Prompting'') method consists of a two-stage approach for tree extraction (c.f. Figure \ref{fig:model_002}). The first model consumes the input image and predicts a set of keypoints corresponding to the root nodes, bifurcations, trifurcations and leaf nodes of the tree. The second stage then extracts patches around each of the keypoints produced by the first stage and recursively decodes the tree structure connecting them together.

\subsection{Keypoint Detection}

The keypoint prediction model uses a simple UNet model to predict small Gaussian ``blobs'' around the locations of the topologically-significant keypoints in the image. During training, the target keypoints grid is computed by placing fixed-size Gaussians (with an application-specific standard deviation) at each node location. During inference, the discrete set of keypoints is extracted from the predicted grid of scalar intensities by using the NMS (non-maximum suppression) algorithm.

\subsection{Recursive Tree Extraction}

\textbf{Connectivity Structure Extraction.} The second stage formulates the tree extraction as a recursive process which consumes both the input image and the set of candidate keypoints. The tree decoding problem is decomposed into a finite set of deterministic recursive steps. During training, each batch is composed of a set of trees from which single randomly-sampled recursive steps in the decoding process are sampled. The model is tasked with selecting for a given parent-query node the corresponding child nodes among the set of candidate keypoints. Since the exact steps of the decoding process are known at training time, this training methodology is fully-supervised and exploration-free. At inference time, each forward pass through the second stage model selects the keypoints which correspond to the child nodes of a specified query node.

\textbf{Image-based Prompt.} In order to indicate to the model the location of the query (parent) node, we compute a ``distance-to-query'' scalar channel $d$ (in world coordinates where the X/Y values go from 0.0 to 1.0 across the image domain) which is then lifted to a ``Fourier feature'' \cite{Tancik2020-td} encoding: $sin(\alpha d)$. We found that this simple encoding was particularly effective in our experiments, a key benefit being that for every position in the image, the local information in this positional encoding is sufficient to determine both the direction towards the query node (from the orientation of the ripples) and the distance to the query node (from the curvature of the ripples). Note that the root node does not have a parent, and for this particular case the image-based prompt is empty. A common fixed scalar $\alpha$ term set to $30.0$ is used for all the positional encodings in this model. 

\textbf{I2TRP Architecture.} The second stage model follows an encoder-decoder architecture. In the encoder pathway, the input image $\boldsymbol{x} \in \mathbb{R}^{H\times W\times C}$ (including the prompt and positional encoding channels) are passed through an 8-layer vision transformer (ViT) \cite{Dosovitskiy2020-fn} to produce a set of global image encodings $\boldsymbol{v_g}$. Around each candidate keypoint location we crop patches $\boldsymbol{p}$ of size 51x51 pixels and pass these through a ResNet \cite{He2015-dg} encoder. The resulting patch-vectors $\boldsymbol{v_p}$ are subsequently fed into a transformer decoder which attends to the global image memory encodings and produces the updated patch vectors $\boldsymbol{v_p'}$. Finally, these updated patch vectors $\boldsymbol{v_p}$ are fed through a two-layer MLP (with a GELU nonlinearity) which predicts the output selection $\boldsymbol{v_s^o}$ and topology $\boldsymbol{v_t^o}$ vectors. Note that both the ViT and ResNets are supplied with the image prompt and positional encoding channels.
\begin{align}
    \boldsymbol{v_g} &= \text{ViT}(\boldsymbol{x}) & \quad
    \boldsymbol{v_p} &= \text{ResNet}(\boldsymbol{p}) \nonumber \\
    \boldsymbol{v_p'} &= \text{TransformerDecoder}(\boldsymbol{v_p}, \boldsymbol{v_g}) & \quad
    \boldsymbol{v_s^o}, \boldsymbol{v_t^o} &= \text{MLP}(\boldsymbol{v_p'}) \nonumber
\end{align}

\textbf{X/Y Positional Encodings.} In addition to the image channel and the dynamic prompt indicating the location of the query node, we also inject X/Y positional information into the patches in the encoder and decoder pathways. While it is possible to add positional encodings to the vector representations in the internal ViT and patch encoder components, this implementation instead represents positional information in the grid external to the model in order to reduce the complexity of the architecture. More specifically, we append a number of static channels to the \emph{input} data which encode the position information. In total there are four channels used for this encoding: two for the absolute X/Y positions (with values between 0.0 and 1.0), and two for sinusoidal lifted variants of these: $sin(\alpha x)$ and $sin(\alpha y)$.

\textbf{Loss Function.} We train the recursive model by combining a pair of loss functions on the output selection and topology vectors $\boldsymbol{v_s^o}$ and $\boldsymbol{v_t^o}$. The selection loss is computed between the predicted selection vector of shape $\mathbb{R}^{P \times 1}$ (where $P$ is the number of patches) and its binary ground-truth vector $\boldsymbol{v_s^g}$ of the same shape. This target selection vector represents the child nodes of the specified query node. Similarly, the topology loss is computed between the predicted topology vector of shape $\mathbb{R}^{P \times 4}$ (note that for the SSA dataset there are no trifurcations and we use topology vectors $\boldsymbol{v_t}$ of shape $\mathbb{R}^{P \times 3}$) and its ground-truth vector $\boldsymbol{v_t^g}$. The target topology vector is a one-hot encoding which represents the different node topologies (the node’s number of children). For both datasets, only root nodes are permitted to have a single child node, and all other nodes either have multiple children or are leaf nodes. Note that root nodes can also have multiple children. To train the topology term we use a cross-entropy loss ($L_{XE}$). We explored using cross-entropy to train the selection term, but found in our experiments that the mean squared error ($L_{MSE}$) loss was more stable. The combined loss $L$ is thus written as:
\begin{align}
    L = \lambda_t \hspace{0.1cm} .\hspace{0.1cm} L_{XE}(\boldsymbol{v_t^o}, \boldsymbol{v_t^g}) +
        \lambda_s \hspace{0.1cm} . \hspace{0.1cm} L_{MSE}(\boldsymbol{v_s^o}, \boldsymbol{v_s^g}) \hspace{0.2cm} , \hspace{0.2cm} \lambda_t = 0.1 , \lambda_s = 1.0
\end{align}

\textbf{Training with Ground-Truth Keypoints.}
We use the ground-truth keypoints to train the second stage I2TRP model instead of using those predicted from the first stage UNet model. In order to mitigate distributional shift, we apply a small spatial Gaussian jitter to the position of the ground-truth keypoints during the training of the second stage. Using the ground truth trees to train the second stage significantly reduces the implementation complexity. At inference time, the candidate keypoints are those predicted by the first-stage UNet, and we find that the recursive model generalises well to the predicted keypoint sets.

\subsection{Stochastic Decoding}

\textbf{Sampling parameters.} Once trained, the tree extraction model can be queried stochastically by interpreting the scalar outputs of the decoder as probabilistic weights (as opposed to simply taking a deterministic argmax). In order to simplify the sampling, we limit this stochasticity to the topology term, and retain the selection as a deterministic prediction at each recursive step. For a given step in the decoding process, the topology head of the decoder predicts a softmax over the node geometry classes for each of the keypoints produced by the first stage. In stochastic mode, we include a ``topology temperature'' term $\gamma$ which adjusts the scalar values of the softmax weights: $w_i = \frac{e^{v_{t,i}^o / \gamma}}{\sum_{j}{e^{v_{t,j}^o / \gamma}}}$. The node topology is then predicted by randomly sampling according to these weights. During inference, the stochastic sampling acquires multiple decodings of the tree, the total number of which is set by the parameter $\mathrm{n}_{\mathrm{dec}}$.

\vspace{0.2cm}

\textbf{Merging the trees.} We implement a merging algorithm which prioritises simplicity and achieves good performance compared to the baseline. This algorithm first produces a matrix which counts the number of times a child and parent keypoint-pair occur in the stochastically generated trees. Once the count matrix is complete, the final merged tree is decoded by starting from the root and recursively selecting the child nodes which most frequently connect to the parent selected at each recursive step.

\section{Data}

\subsection{Volumetrically Rendered Meshes (VRM)}\label{sec:vrm}

 The first dataset which we utilize in this paper is generated by rendering a large set of 3D coronary artery tree meshes derived from CTA (computed tomography angiography). Our intent is to create a large-scale ``semi-synthetic'' dataset using real anatomical data which replicates the geometric complexity of the tree connectivity structure extraction problem as seen in real-world modalities such as projected 2D X-ray angiography. Note that the dataset is acquired from geographically diverse regions, and is representative of clinical populations.

\begin{figure}[h]
    \centering
    \subfloat[]{
        \label{fig:data_tree_37}
        \includegraphics[width=0.3\textwidth]{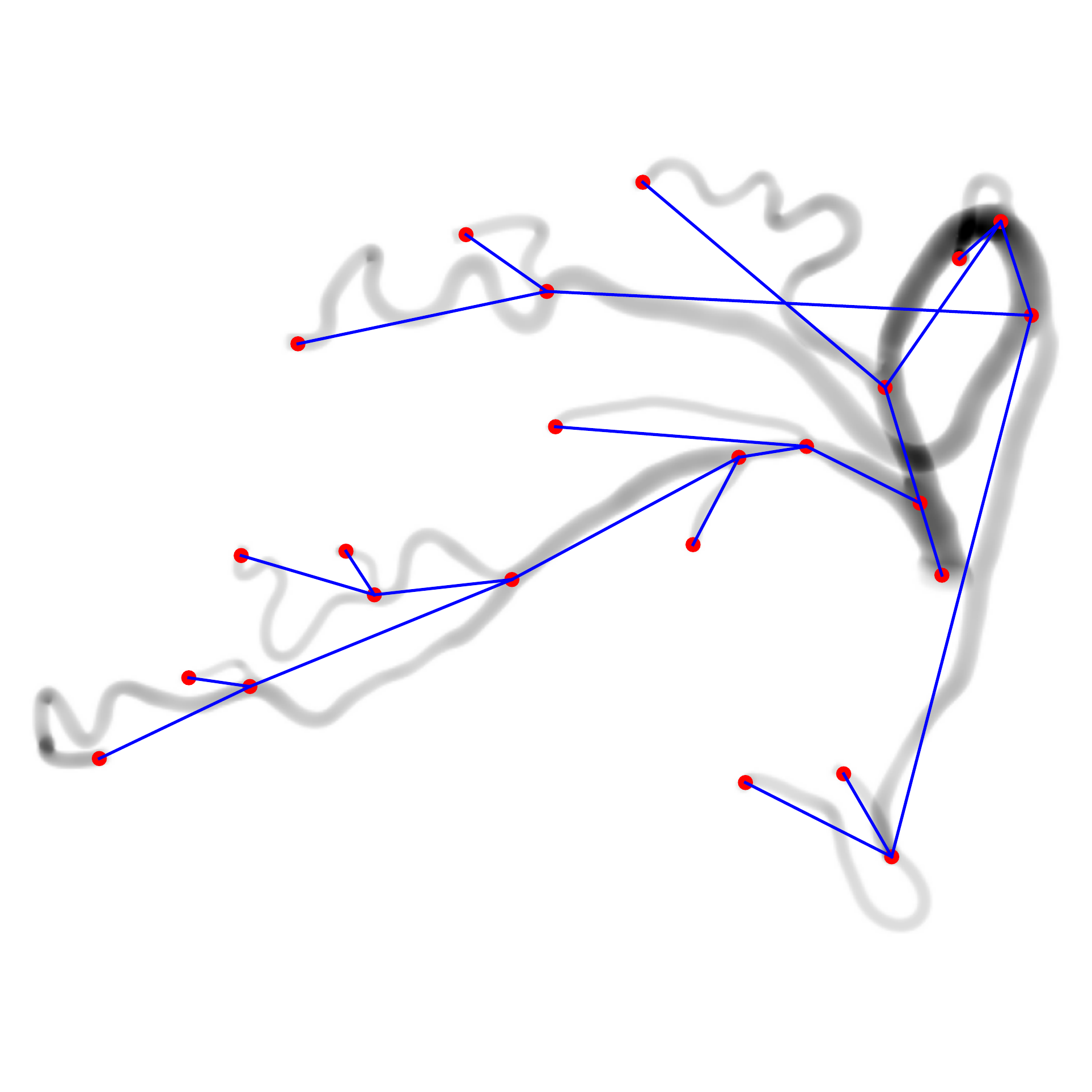}
    }
    \hspace{0.1cm}
    \subfloat[]{
        \label{fig:data_tree_90}
        \includegraphics[width=0.3\textwidth]{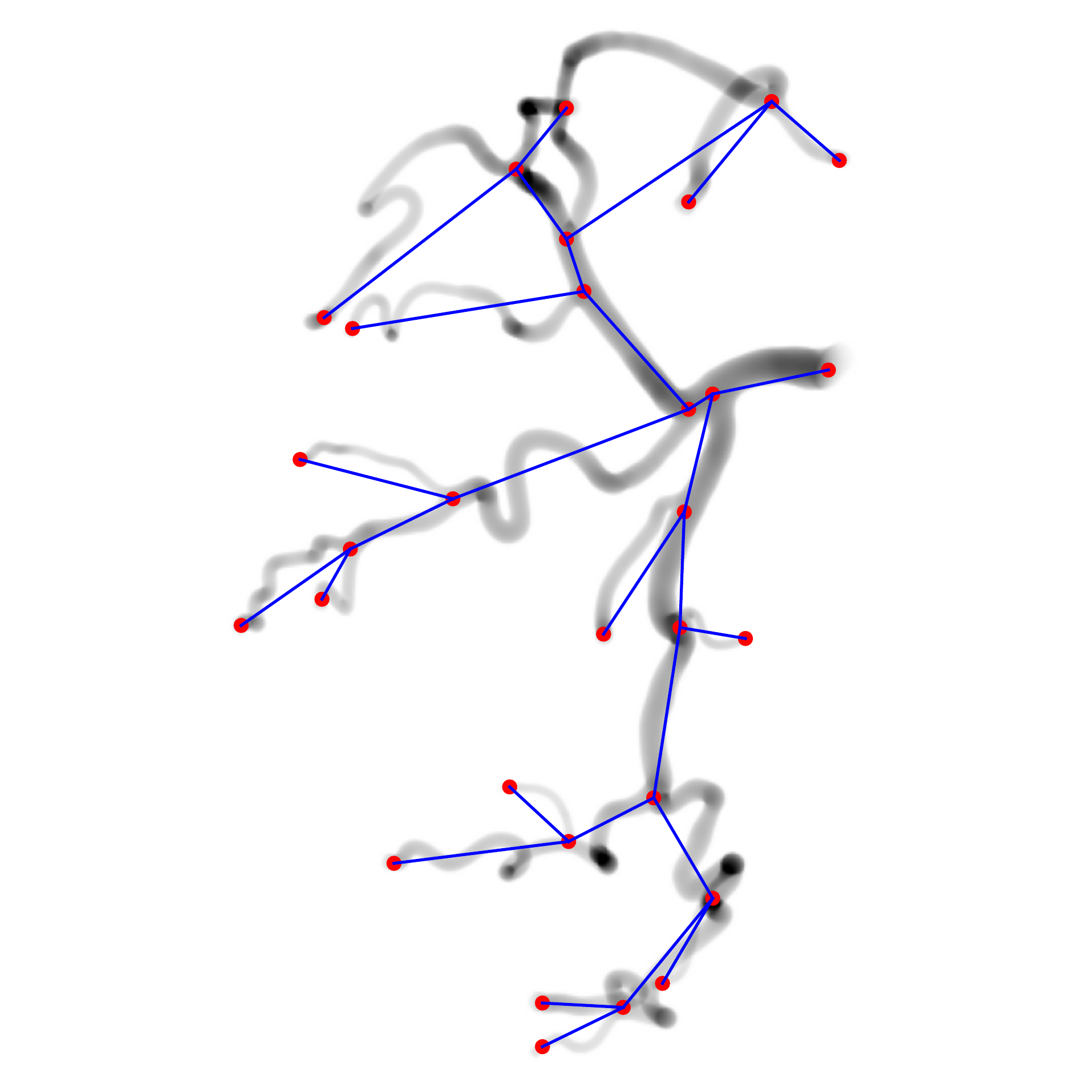}
    }
    \hspace{0.1cm}
    \subfloat[]{
        \label{fig:data_tree_24}
        \includegraphics[width=0.3\textwidth]{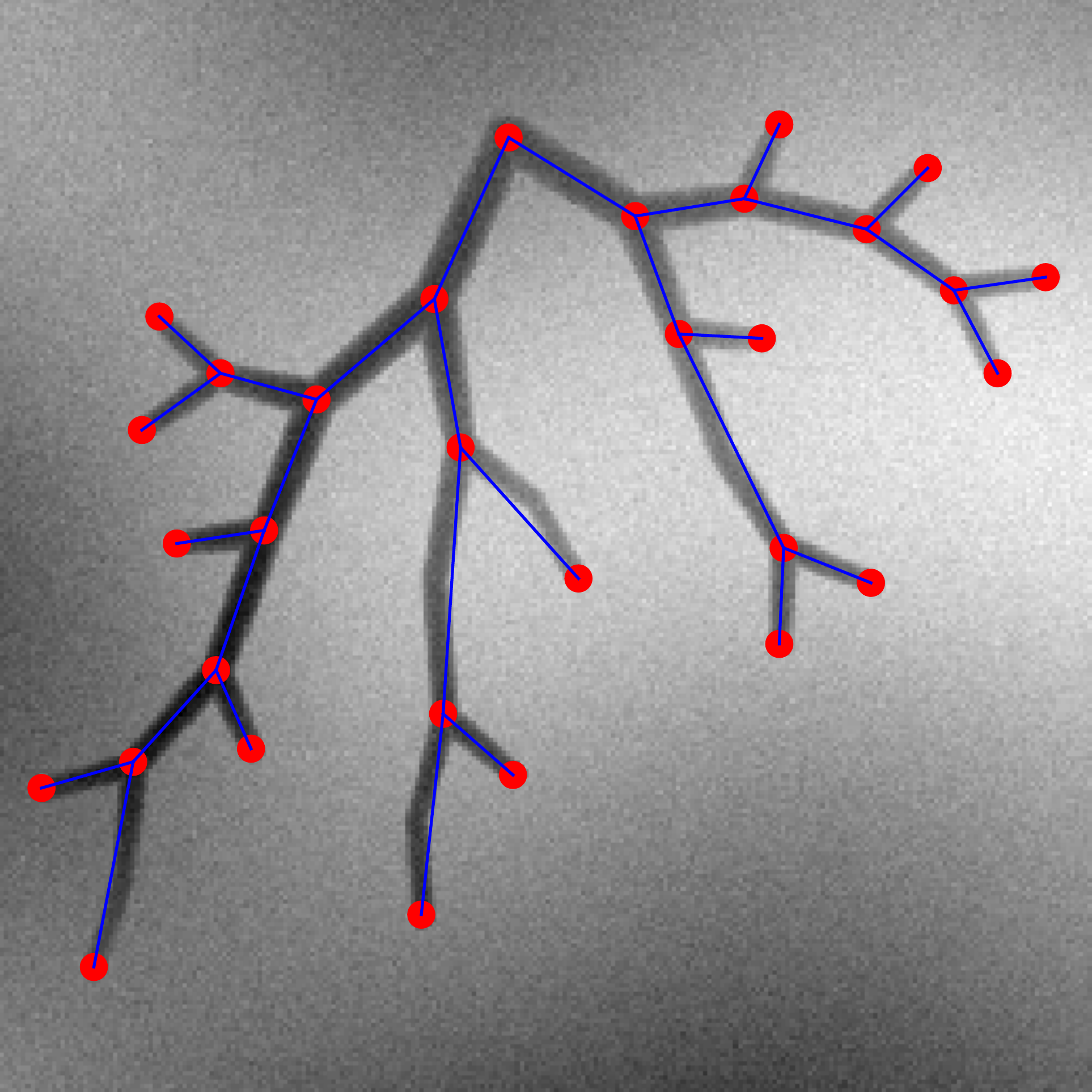}
    }
    \caption{Examples with the ground-truth tree connectivity structure overlay. (a) and (b): from the VRM dataset. (c): from the SSA dataset}
    \label{fig:ssa_examples}
\end{figure}

The VRM dataset is generated from 9,845 left and right coronary tree meshes. For each tree we generate volumetric projections along five different view angles. These 2D projections are then split into three subsets: train (39,420 views), test (8,855 views) and validation (950 views). Note that we ensure trees from the same patient are not mixed between different subsets. The trees in the VRM dataset include root, leaf, bifurcation and trifurcation nodes. Since the trees are generated from real-world data, the relative frequencies of these reflect the true statistics of coronary anatomy. Since the 3D meshes are clipped beyond the radius of 0.25mm, small secondary vessel structures at finer resolutions are not present in this representation. For each view, we render 500x500 pixel grids by randomly sampling perspective projections and tracing rays through the scene. For every pixel in these grids, we set the intensity according to the corresponding ray traversal distance through the coronary artery mesh (c.f. Fig. \ref{fig:data_tree_37} and \ref{fig:data_tree_90}).

\subsection{Simple Synthetic Angiography (SSA)}\label{sec:ssa}

In addition to the volumetrically rendered meshes, we generate a simpler 2D synthetic dataset, composed of 15,248 train and 3,812 test trees, on which both the proposed and baseline models are trained and evaluated. This dataset, which will be publicly released, is intended to both facilitate reproducibility of our work and support future development of learning-based tree extraction algorithms.

The dataset is generated in two steps: the first step produces the tree geometry, and the second renders the corresponding image. The first step models the angiogenesis process using a simple force-based simulation which iteratively grows a tree composed of a set of nodes and edges. In order to assign the vessel diameter to each location along the centerline, we use Murray's law \cite{Murray1926-hj}. Note that the SSA dataset does not contain trifurcations. The second step consumes the tree geometry and produces the noisy rendered image. In order to emulate vessel-like appearance, we model each vessel as a 3D cylinder (aligned in the same plane) filled with constant-density contrast. For each tree, we create a grid of 250x250 pixels and trace rays orthographically through the scene. As a final step in the rendering process, we add multi-scale Perlin noise (c.f. Fig. \ref{fig:data_tree_24}).

\section{Baseline}

Our model is evaluated against a baseline inspired by previous works which follow the minimum cost path approach \cite{Cohen1997-si,Benmansour2011-to,Li2006-ct}. The extraction of curvilinear structures in image data typically involves optimizing a candidate path $\chi$ according to a function $f: \chi \rightarrow \mathbb{R}$. Methods which leverage minimal path techniques include those which define cost or potential maps derived from image intensities \cite{Cohen1997-si}, vesselness \cite{Metz2009-ld} and medialness \cite{Gulsun2008-fz} filters or distance transforms of segmentation masks \cite{Frimmel2004-ea}. Image-domain cost maps $f(I)$ assign for every discrete pixel location a particular cost, allowing integration along a sequence of pixels to recover the path score. The choice of cost map $f(I)$ affects the resulting minimum-cost path. 

In our baseline implementation, the image of the tree is passed through a UNet \cite{Ronneberger2015-tg} which predicts the set of root and leaf nodes, in addition to a segmentation mask $S$. A distance transform $D_{int}(S)$ is then applied inside the predicted mask which represents for each pixel the distance to the segmentation edge. The cost map $C$ is computed such that on the mask's interior $C_{int} = m - D_{int}(S)$, and on its exterior $C_{ext} = m + 1$, where $m = \max({D_{int}(S)})$. Minimal-cost paths are then traced accordingly from each leaf node to the root (using Dijkstra's algorithm). Finally, these paths are merged to form the centerline tree, from which the discrete tree connectivity structure is extracted.

\section{Training Details}

We trained both the proposed and baseline models using the AdamW \cite{Loshchilov2017-ra} optimizer implemented in PyTorch \cite{Paszke2019-th}. For training, the learning rate scheduler performs a short linear ramp up followed by an exponential learning rate decay, with different peak learning rate and decay parameters. All experiments were run on a single NVIDIA V100 32GB GPU, with a weight decay of 1e-3, using translate, scale and rotate augmentations.

\subsection{First stage: Keypoint model}

On both the SSA and VRM datasets, the first stage UNet model is trained using a batch size of 16. The grid containing the target gaussian blobs is partitioned by thresholding at 0.5, and the MSE foreground and background terms are reweighted by multiplying by 0.7 and 0.3 respectively. On the SSA dataset, the peak learning rate after the linear ramp up is 1e-3, versus 3e-4 for the VRM dataset. The UNet model is trained for 21k steps on the SSA dataset, with the learning rate decaying by a factor 10 every 10k steps, and for 66k steps on the VRM dataset, with the learning rate decaying by a factor 10 every 20k steps.

\paragraph{\textbf{Augmentation.}} Note that since the scale augmentation is used while training this stage, we take care to compute the target keypoints grid dynamically instead of applying augmentations to the target image. This ensures that the generated gaussian blobs are of fixed pixel dimensions.

\subsection{Second stage: Recursive Model}

On the SSA dataset the second stage model uses a transformer decoder with 12 layers and ViT with a patch size of 25. On the VRM dataset a decoder with 21 layers is used and a patch size of 40. On both datasets we use a peak learning rate of 1e-4 and a batch size of 144. On the SSA dataset the learning rate is decayed by a factor 10 every 10k steps, and every 50k steps on the VRM dataset. In all our experiments we use transformers with eight heads and a per-head feature dimension of size 64.

\section{Results}

\subsection{Evaluation Metrics}

\begin{sloppypar}
In order to compare the performance of our proposed I2TRP model against the baseline, we use leverage two point cloud distance metrics: the Chamfer distance $ \mathrm{CD}(S_a, S_b) = \frac{1}{|S_a|}\sum_{x \in S_a}{\min_{y \in S_b} ||x - y||^2_2} + \frac{1}{|S_b|}\sum_{y \in S_b}{\min_{x \in S_a} ||x - y||^2_2}$ and the Hausdorff distance $\mathrm{HD(S_a, S_b)} = \max\{\hspace{0.1cm}\max_{x \in S_a}\min_{y \in S_b} ||x - y||_2, \hspace{0.1cm} \max_{y \in S_b}\min_{x \in S_a} ||x - y||_2\hspace{0.1cm}\}$. While the objective here is to quantitatively compare 2D predicted \emph{tree connectivity structures} against their ground-truth targets, we found in practice that these set-to-set distance metrics are effective measurement tools which are both sensitive to small errors and correlate well with qualitative assessment. In order to compute these metrics between a predicted and target tree, we linearly sample 100 points down each edge, and then compute for each case the distance metrics between the two point sets. This metric is then averaged over the test (or validation) set to obtain the final result.
\end{sloppypar}

\subsection{SSA Test Set}

The results on the SSA dataset indicate that our proposed I2TRP model outperforms the baseline (c.f. Table \ref{table:vrm_ssa_test_set}). We emphasize, however, that this dataset is intended to be simple by design, and that the absolute difference between the models on these two metrics is fairly marginal considering these distances are measured in \emph{pixels} (and $\mathrm{pixels}^2$).

\begin{figure}[h]
    \centering
    \subfloat[]{
        \label{fig:ssa_baseline}
        \includegraphics[width=0.23\textwidth]{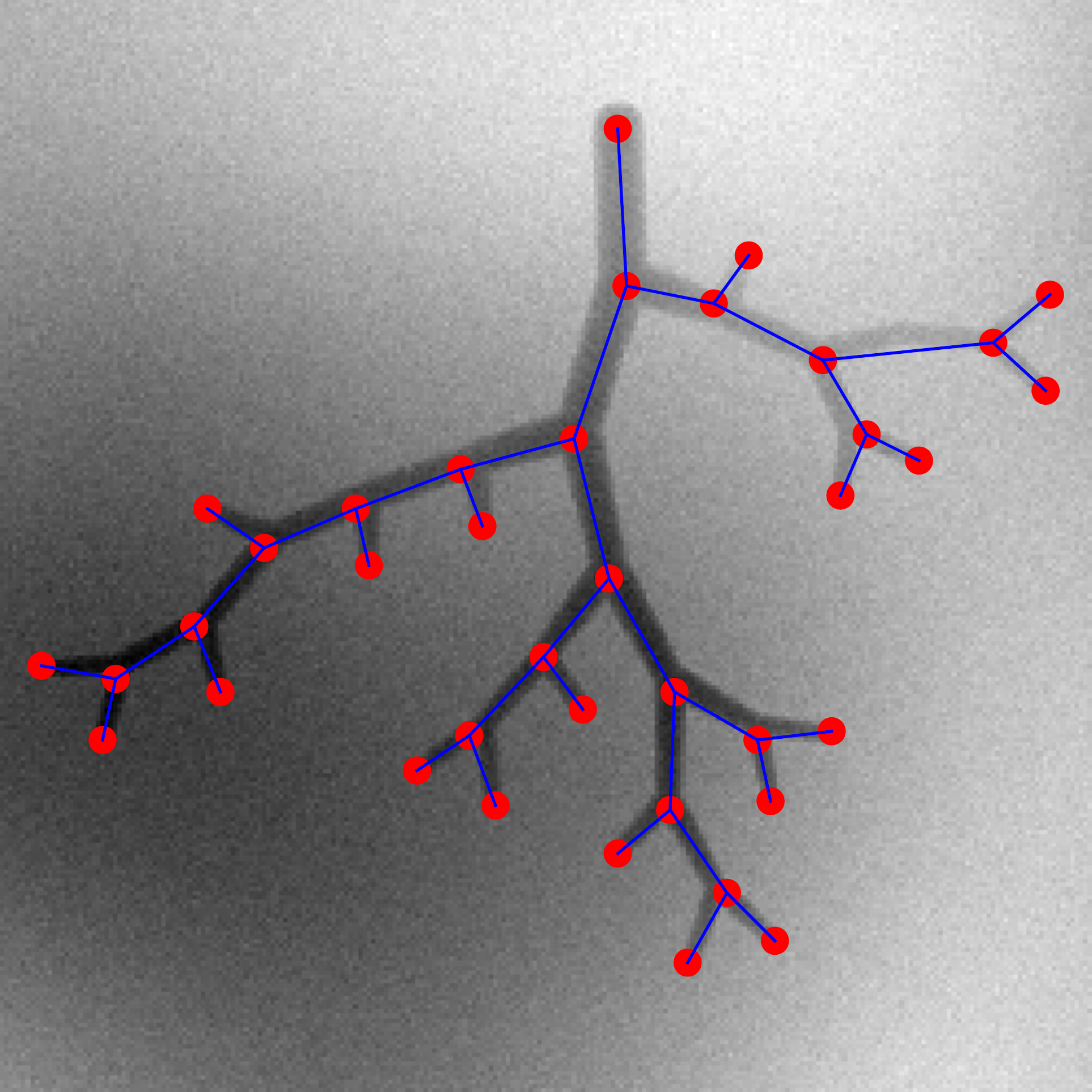}
    }\hspace{0.3cm}
    \subfloat[]{
        \label{fig:ssa_proposed}
        \includegraphics[width=0.23\textwidth]{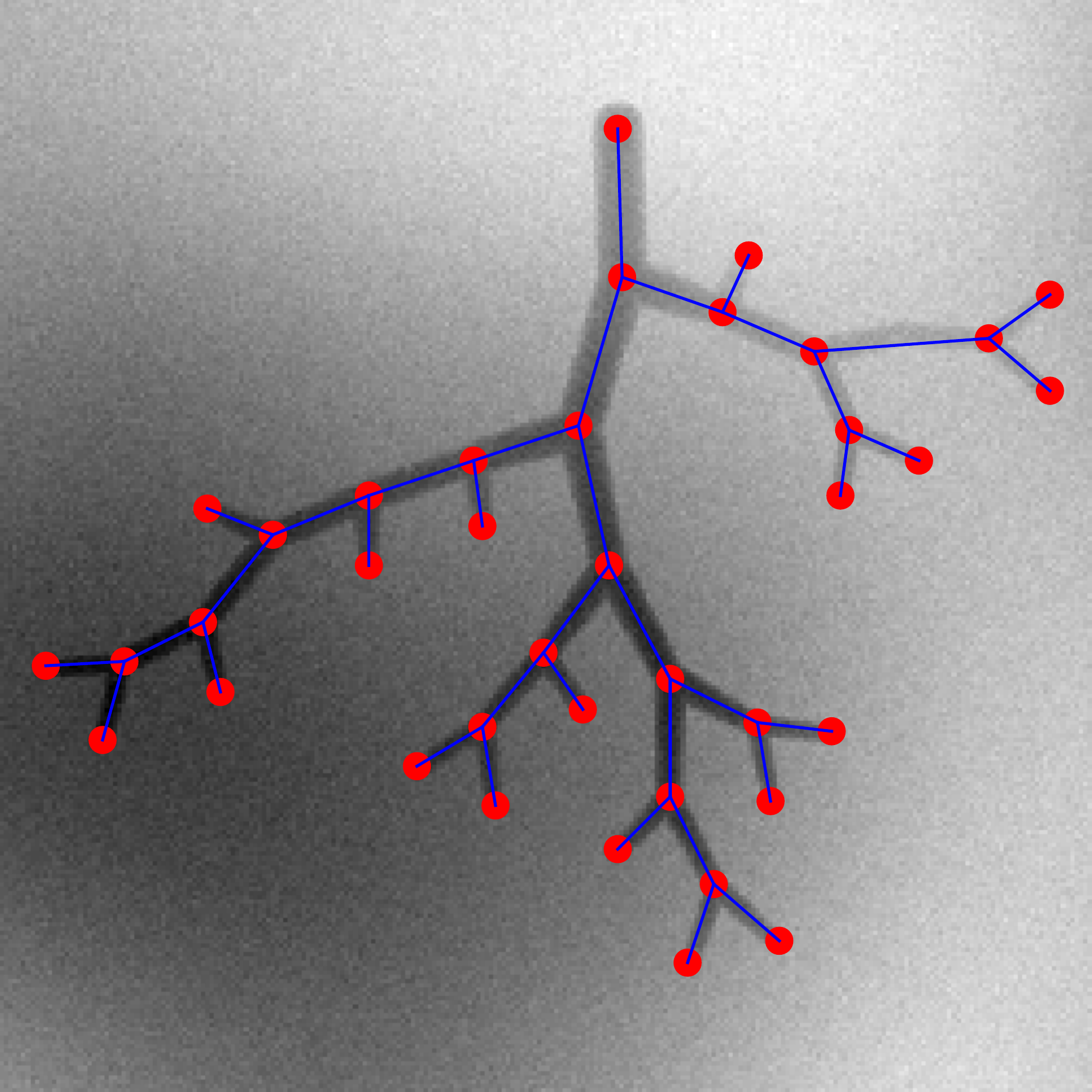}
    }
    \caption{Overlay of the tree connectivity structure predicted by the baseline (a) and proposed model (b) for one
           example in the SSA test set}
\end{figure}

For the inference on the SSA dataset, we use the sampling parameters $\mathrm{n_{dec}}=10$ and $\gamma=1.0$. While these parameters are possibly suboptimal, we chose to not over optimize and this simple data, and leave the sensitivity analysis of different sampling parameters to the following section on the VRM data. Qualitatively, the tree connectivity structures predicted by both the baseline and I2TRP models are highly accurate on the SSA dataset (c.f. Figures \ref{fig:ssa_baseline} and \ref{fig:ssa_proposed}). 

\subsection{VRM Validation and Test Sets}

On the VRM validation set, we seek to explore the sensitivity of the $\mathrm{n}_\mathrm{dec}$ and $\gamma$ sampling parameters in order to achieve improved performance on the test set.

\begin{figure}[h]
    \centering
    \subfloat[]{
        \label{fig:vrm_h_varing_temp}
        \includegraphics[width=0.32\textwidth]{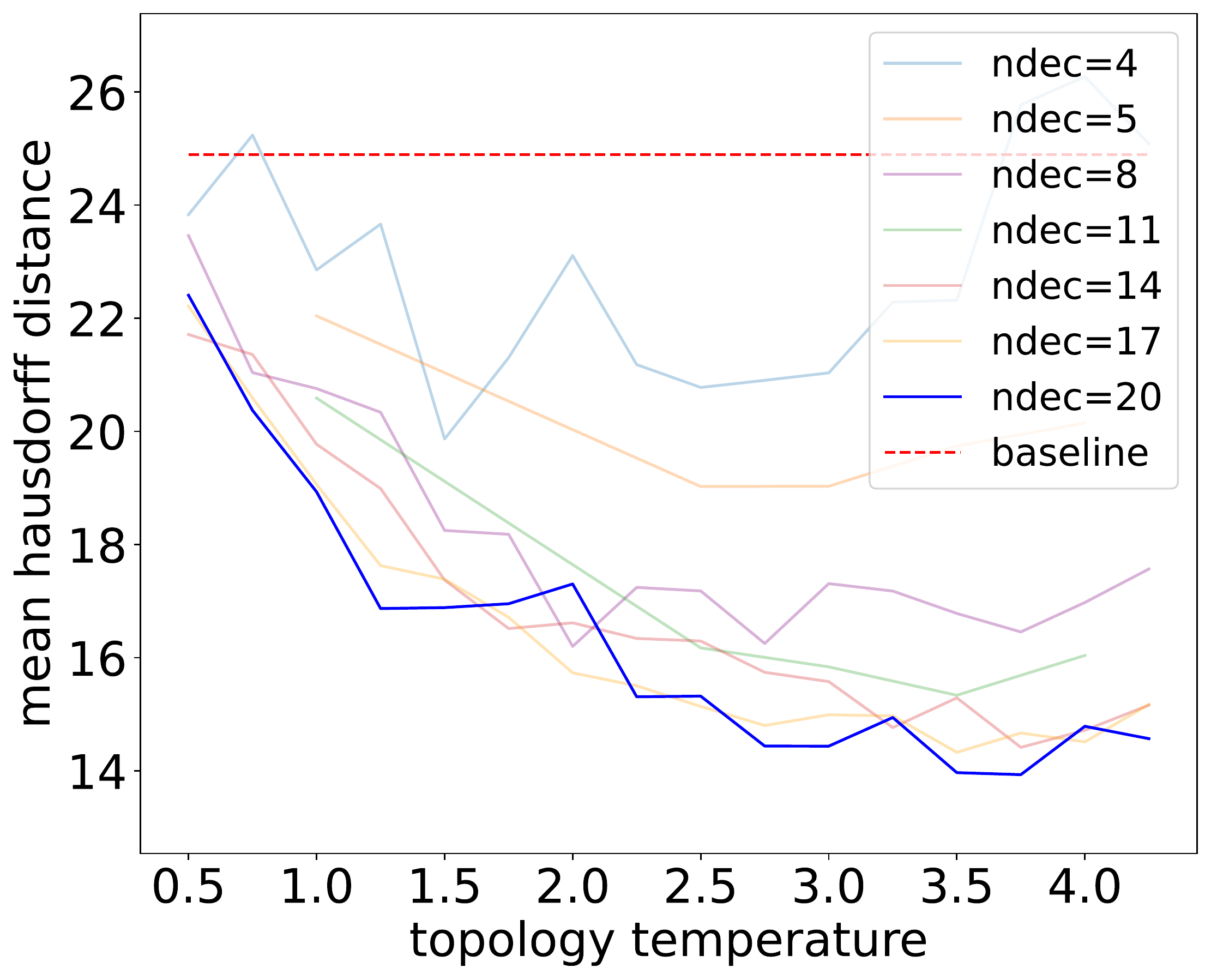}
    }
    \subfloat[]{
        \label{fig:vrm_h_varying_ndec}
        \includegraphics[width=0.32\textwidth]{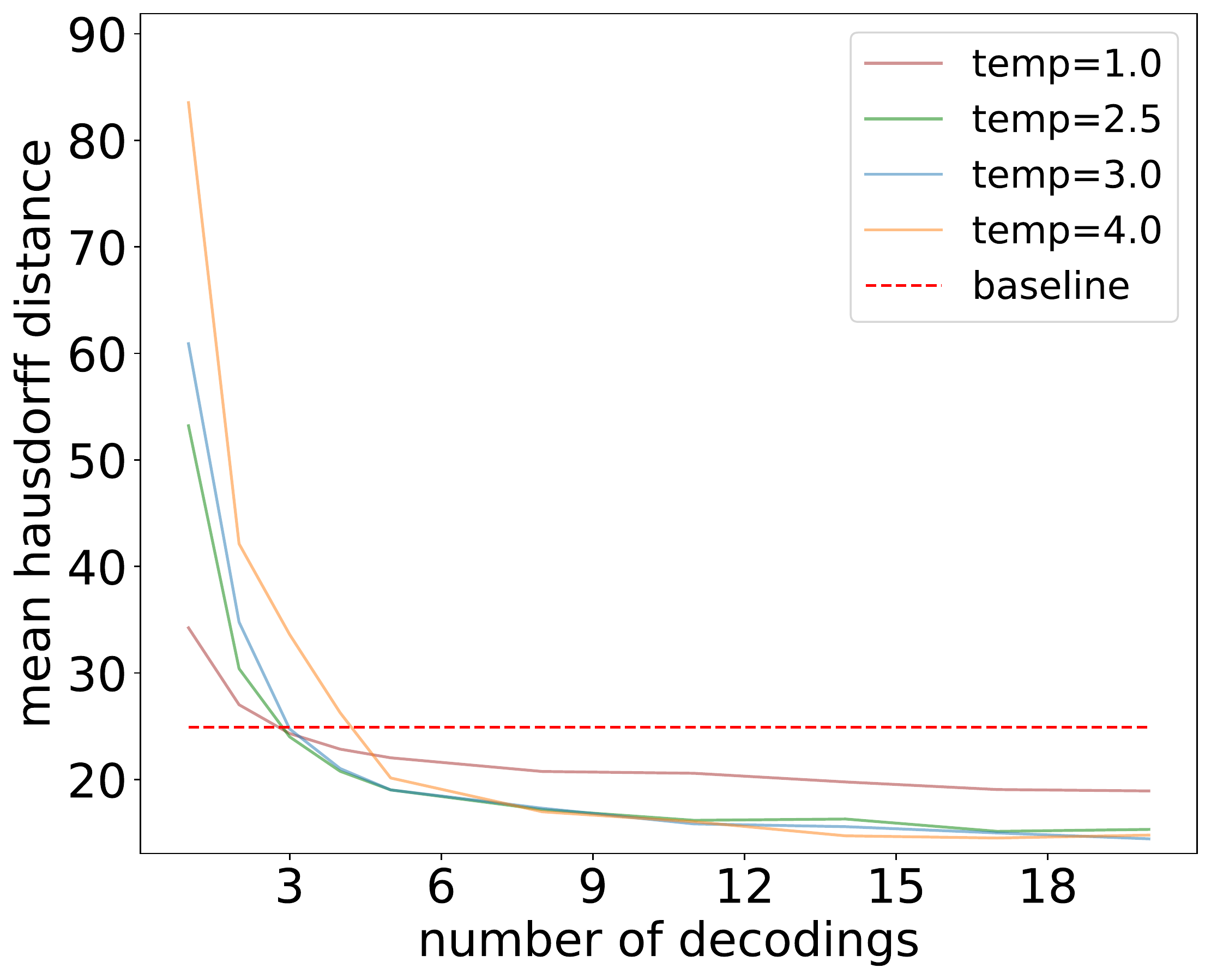}
    }
    \subfloat[]{
        \label{fig:vrm_c_varing_temp}
        \includegraphics[width=0.335\textwidth]{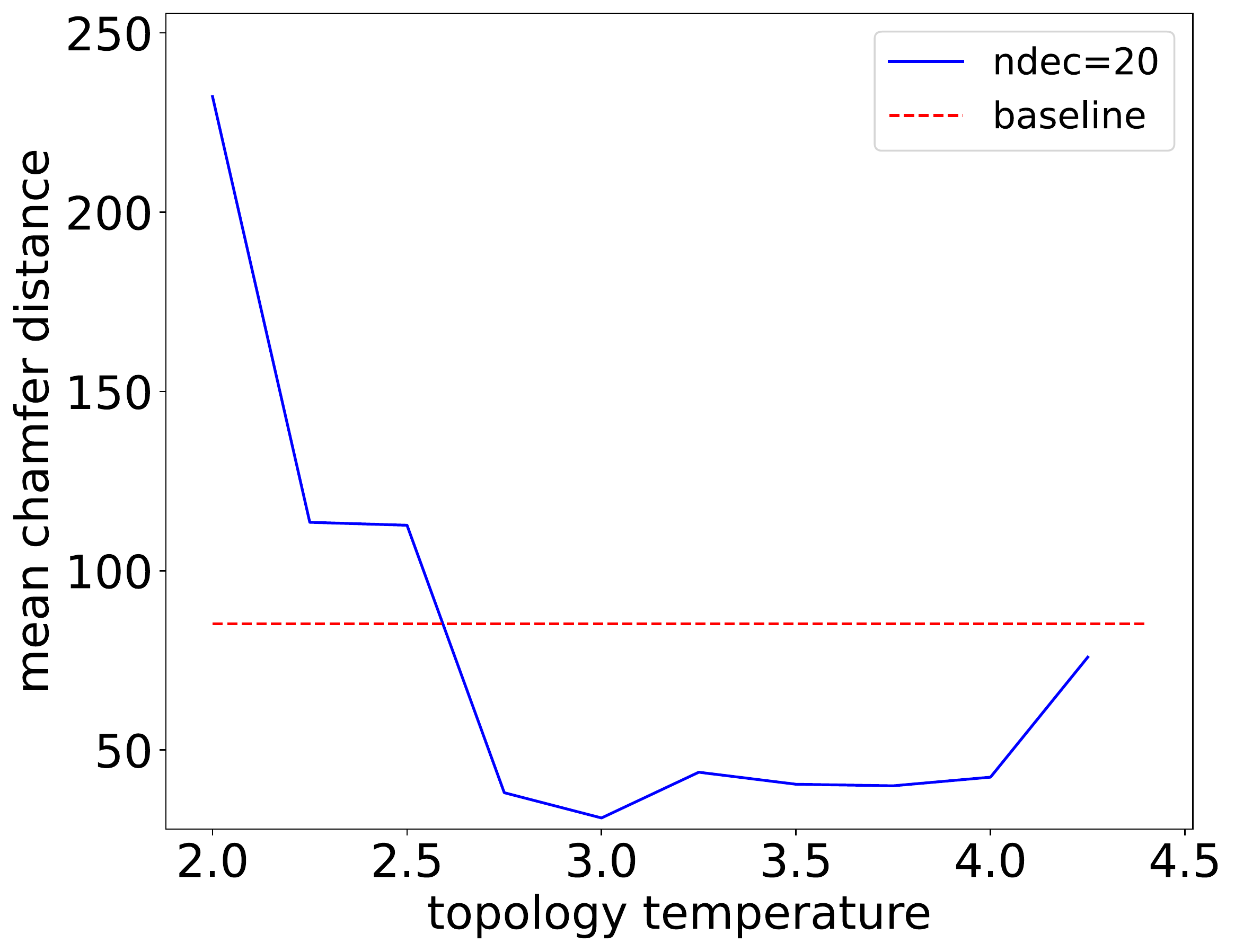}
    }
    \caption{Results of the proposed model on the VRM validation set. (a): mean Hausdorff distances for multiple values of $\mathrm{n}_\mathrm{dec}$ as a function of the temperature values. (b): mean Hausdorff distances for multiple values of $\gamma$ (topology temperature) as a function of $\mathrm{n}_\mathrm{dec}$. (c): mean Chamfer distance obtained on the validation set for $\mathrm{n}_\mathrm{dec}=20$ as a function of the temperature values.}
\end{figure}

\textbf{Sensitivity of the number of stochastic samples}. The first sampling parameter which we explore is the $\mathrm{n}_\mathrm{dec}$ term, which corresponds to the number of tree connectivity structures which are stochastically sampled before being merged into the final tree. In Figure \ref{fig:vrm_h_varying_ndec} and Table \ref{table:vrm_val_set} we can observe that for a variety of different $\gamma$ (topology temperature) terms, the quality of the merged tree consistently improves as the number of stochastic samples increases.

\begin{figure}[h]
    \centering
    \subfloat[]{
        \label{fig:vrm_tree_rendering}
        \includegraphics[width=0.33\hsize]{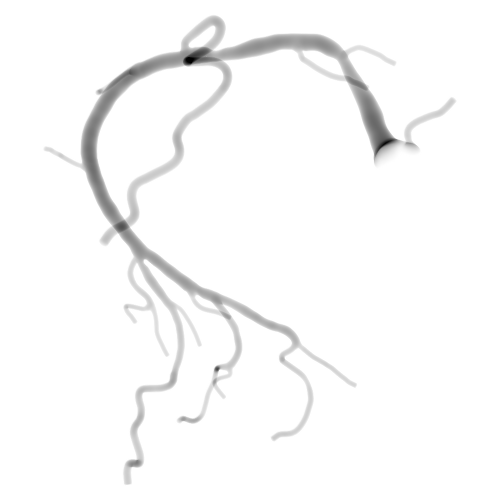}
    }
    \subfloat[]{
        \label{fig:vrm_tree_proposed}
        \includegraphics[width=0.33\textwidth]{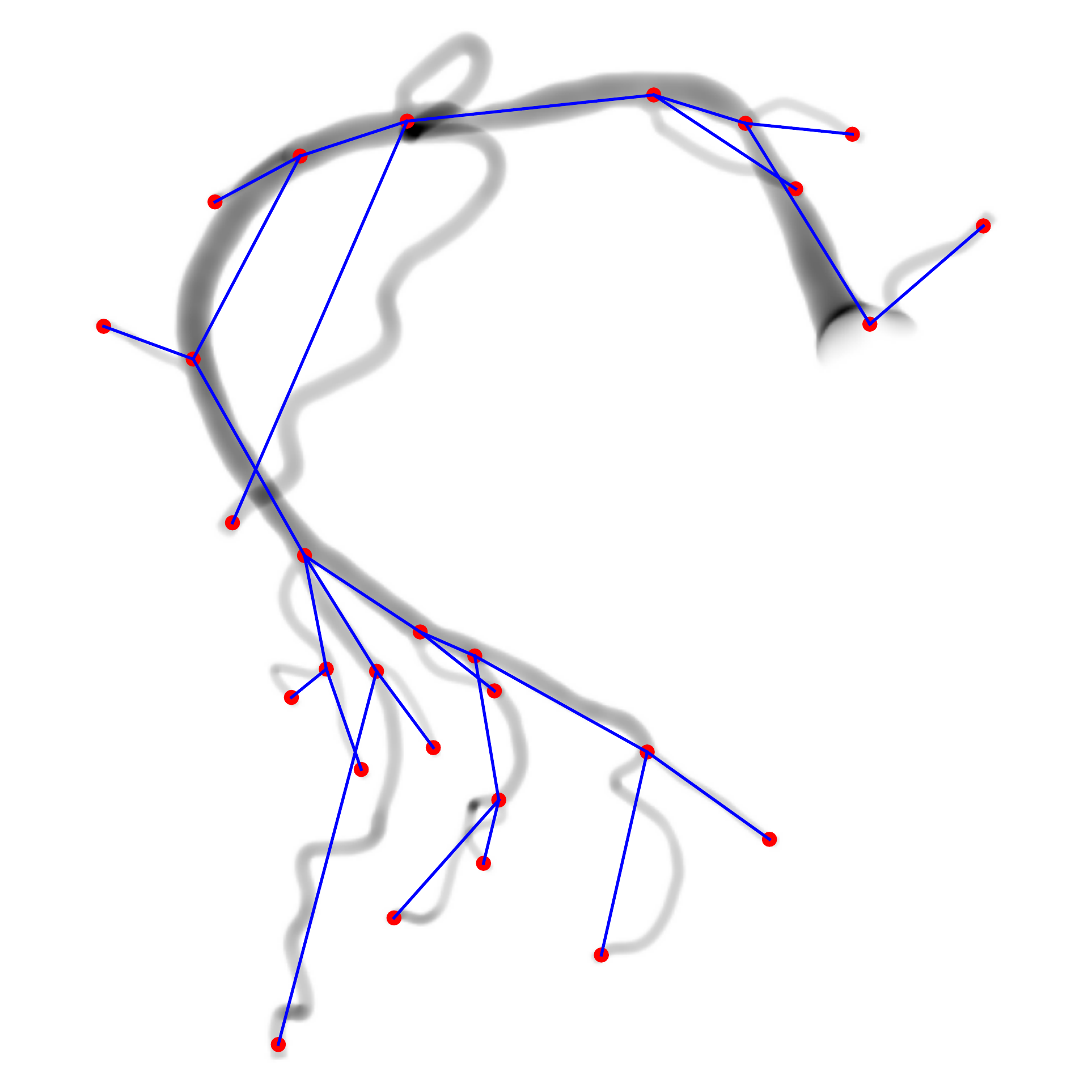}
    }
    \subfloat[]{
        \label{fig:vrm_tree_baseline}
        \includegraphics[width=0.33\textwidth]{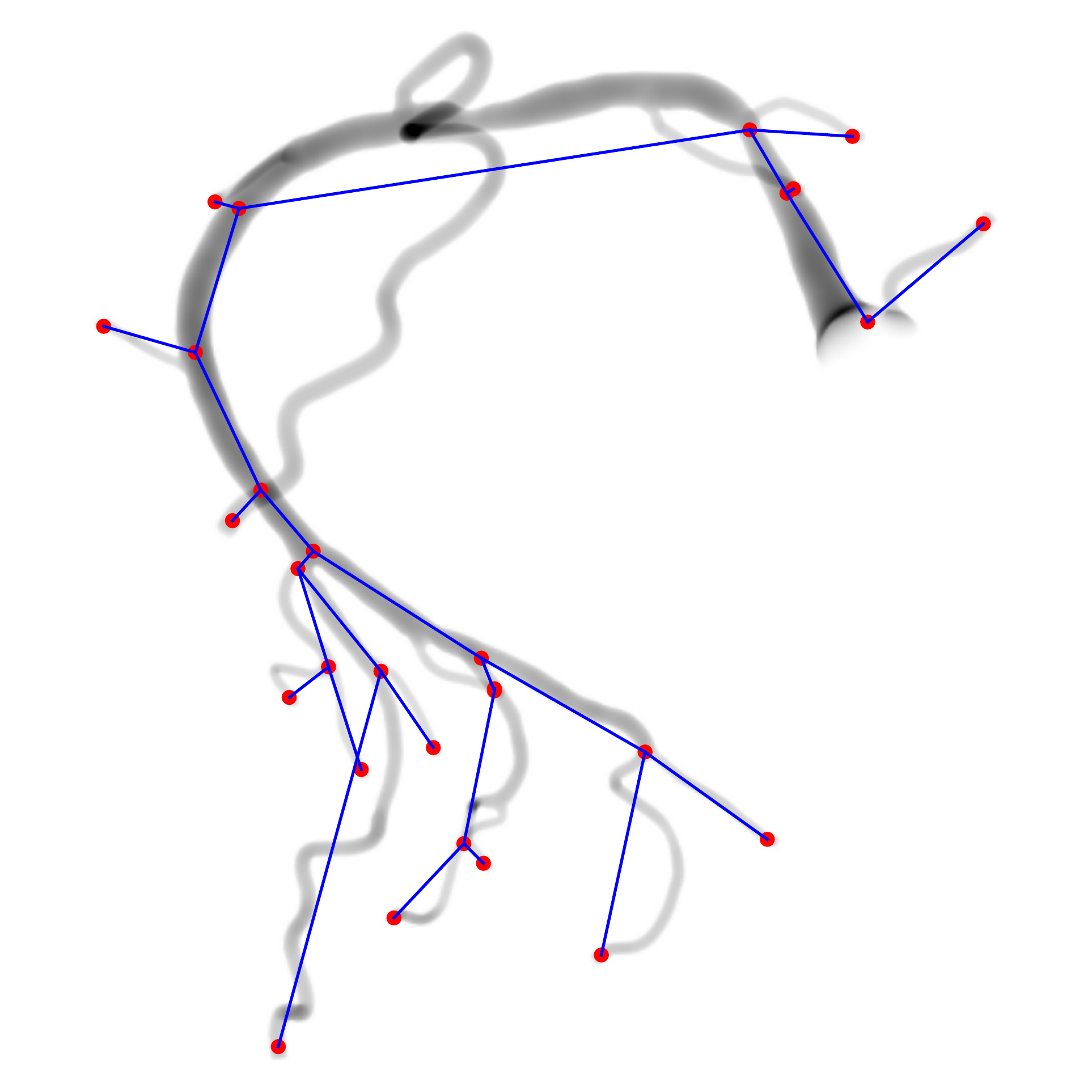}
    }
    \caption{Predicted tree connectivity structures on an example from the VRM validation set. (a): input image, (b): I2TRP prediction, (c): baseline prediction. Of particular note here is the shortcut problem demonstrated by the baseline in Figure (c), observed at the centre and the top right. This problem does not occur in the proposed I2TRP model (Figure (b)) since it directly extracts the tree connectivity structure.}
\end{figure}

\textbf{Sensitivity of the temperature term}. The second sampling parameter which we explore is the $\gamma$ term, which corresponds to the softmax temperature used to sample the categorical topology classification. In Figures \ref{fig:vrm_h_varing_temp} and \ref{fig:vrm_c_varing_temp} and Table \ref{table:vrm_val_set} we observe that for for different values of $\mathrm{n}_\mathrm{dec}$, we can select a temperature term which reaches the minimum Chamfer distance on the VRM validation set.

Analysing the sensitivity of these two sampling parameters across a range of values suggests that the possibly best performance on the VRM data is achieved for $\mathrm{n}_\mathrm{dec} = 20$ and $\gamma=3.0$. For these parameters, the proposed model reaches a Chamfer distance of 31.06 $\mathrm{pixels}^2$ on the VRM validation set (vs 85.20 $\mathrm{pixels}^2$ for the baseline model) . Note that the I2TRP model's $\mathrm{HD}$ of 14.44 pixels (vs 24.89 pixels for the baseline) is close to the optimum according to these results, but that a slightly better $\mathrm{HD}$ value of 13.97 is reached for a value of 3.5 for $\gamma$.

\begin{table}[h]
\begin{minipage}[h]{0.55\linewidth}
\begin{adjustbox}{width=1.0\linewidth, center}
\begin{tabular}{ c|c|c|c|c|c|c||c } 
\hline
$\mathrm{n}_\mathrm{dec} = 20$ \\
\hline
$\gamma$ & 1.5 & 2.0 & 2.5 & 3.0 & 3.5 & 4.0 & baseline \\
\hline
\multirow{1}{4em}{$\mathrm{HD}$} & 16.89 & 17.30 & 15.32 & 14.44 & \textbf{13.97} & 14.79 & 24.89 \\
\multirow{1}{4em}{CD} & 146.6 & 232.2 & 112.7 & \textbf{31.06} & 40.46 & 42.44 & 85.20 \\
\hline
\hline
$\gamma=3.0$ \\
\hline
$\mathrm{n}_\mathrm{dec}$ & 5 & 8 & 11 & 14 & 17 & 20 & baseline\\
\hline
\multirow{1}{4em}{$\mathrm{HD}$} & 19.03 & 17.31 & 15.84 & 15.58 & 14.99 & \textbf{14.44} & 24.89 \\
\multirow{1}{4em}{CD} & 277.4 & 177.5 & 91.22 & 160.8 & 57.01 & \textbf{31.06} & 85.20   \\
\hline
\end{tabular}
\end{adjustbox}
\captionsetup{font=small}
\vspace{0.3cm}
\caption{VRM validation set sensitivity analysis}
\label{table:vrm_val_set}
\end{minipage}
\hspace{0.06\linewidth}
\begin{minipage}[h]{0.32\linewidth}
\begin{adjustbox}{width=1.34\linewidth, center}
\begin{tabular}{ c|c|c }
\hline
\textbf{SSA Test Set} \\
\hline
Model & $\mathrm{HD}$ & CD \\
\hline
\multirow{1}{4em}{Baseline} & 3.55 & 2.60 \\
\multirow{1}{4em}{I2TRP} & $\boldsymbol{1.36 \pm 0.06}$ & $\boldsymbol{0.67 \pm 0.10}$ \\
\hline
\hline
\textbf{VRM Test Set} \\
\hline
Model & $\mathrm{HD}$ & CD \\
\hline
\multirow{1}{4em}{Baseline} & 25.26 & 92.16 \\
\multirow{1}{4em}{I2TRP} & \textbf{15.02} & \textbf{71.73} \\
\hline
\end{tabular}
\end{adjustbox}
\captionsetup{font=small}
\vspace{0.2cm}
\caption{Test sets results}
\label{table:vrm_ssa_test_set}
\end{minipage}

\end{table}

Finally, when evaluating both models on the VRM test set with these fixed sampling parameters, we observe quantitatively that the proposed model outperforms the baseline (c.f. Table \ref{table:vrm_ssa_test_set}). A qualitative evaluation of the decoded tree structures (c.f. Figures \ref{fig:vrm_tree_rendering}, \ref{fig:vrm_tree_proposed}, and \ref{fig:vrm_tree_baseline}) further suggests that our proposed I2TRP model does not suffer from the baseline's shortcut issue, and can correctly handle branches overlapping in the projection (c.f. centre and top right of Figure \ref{fig:vrm_tree_baseline}).

\section{Conclusion \& Discussion}

We presented a novel algorithm to extract tree connectivity structure from images. We described two large-scale 2D synthetic datasets, one of which is generated using real 3D coronary artery meshes from a clinically representative patient population. Our model achieves favorable performance compared to a minimal-path baseline. We report the influence of two sampling parameters on a validation set. We expect that solving the connectivity structure problem in modalities such as projective X-ray Angiography will significantly reduce the complexity of subsequently extracting the full curvilinear centerline tree. Since high quality expert annotations are expensive to produce at scale for tree-structured data, our work here is currently limited to synthetic images. Bridging the gap between real and synthetic data is rapidly becoming more tractable thanks to diffusion models, and we intend to further explore combining these with our method. Furthermore, while the model proposed in this paper is applied only to 2D images, we are confident that the choice of model architecture will scale well to 3D modalities such as CT angiography.

\section{Acknowledgements}

This research was funded by HeartFlow, Inc.; James Batten was supported by the UKRI CDT in AI for Healthcare \url{http://ai4health.io} (Grant No. EP/S023283/1).

%
%
\bibliography{mybib} 
\bibliographystyle{splncs04}
%

\end{document}